\documentclass[letterpaper, 10 pt, conference]{ieeeconf}  

\IEEEoverridecommandlockouts                              

\usepackage{amsmath}
\usepackage{graphicx}
\usepackage{amsfonts}
\usepackage{tabularx}
\usepackage{booktabs}
\usepackage{ragged2e}
\usepackage{multirow}
\usepackage{xcolor}
\usepackage{comment}

\title{\LARGE \bf
Speech2Grasp: Data-Efficient Transfer of Text-Conditioned Grasp Detection to Speech in Humanoid Robots
}

\author{
Hung Nguyen$^{1,*}$,
Kim Nhat Minh Nguyen$^{2,*}$,
Van Duc Vu$^{2}$,
Van-Danh Le$^{3}$,
Hoang Huy Le$^{2}$,
\\
Dinh Tuan Nguyen$^{2}$,
Pham Tuyen Le$^{2}$,
Van-Truong Nguyen$^{2,4}$,
Quan Nguyen$^{2,5}$%
\thanks{$^{*}$Equal contribution. $^{1}$University of California San Diego, $^{2}$VinMotion, $^{3}$Hanoi University of Science and Technology, $^{4}$VinUniversity, $^{5}$University of Southern California}
}

\begin{document}

\maketitle
\thispagestyle{empty}
\pagestyle{empty}

\begin{abstract}
Humanoid robots increasingly require multi-modal understanding for natural interaction with humans. Despite the prominence of vision-language models, they generally assume textual rather than the more natural speech inputs. In this paper, we investigate whether a well-established text-conditioned model can be transferred to speech in a data-efficient manner. Using ALBEF as a case study, we conduct diagnostic analyses showing that a lightweight MLP-based projector effectively adapts it to speech, while preserving semantic discrimination and robustness. Motivated by these findings, we introduce Speech2Grasp, a framework for data-efficient transfer of text-conditioned grasp detection to speech. Real-world humanoid robot experiments show that Speech2Grasp outperforms cascaded ASR-based pipeline, while reducing inference latency. Our findings suggest a practical paradigm for extending established text-conditioned systems to speech. 
\end{abstract}

\section{Introduction}

\begin{figure}[thbp]
    \centering
    \includegraphics[width=\linewidth]{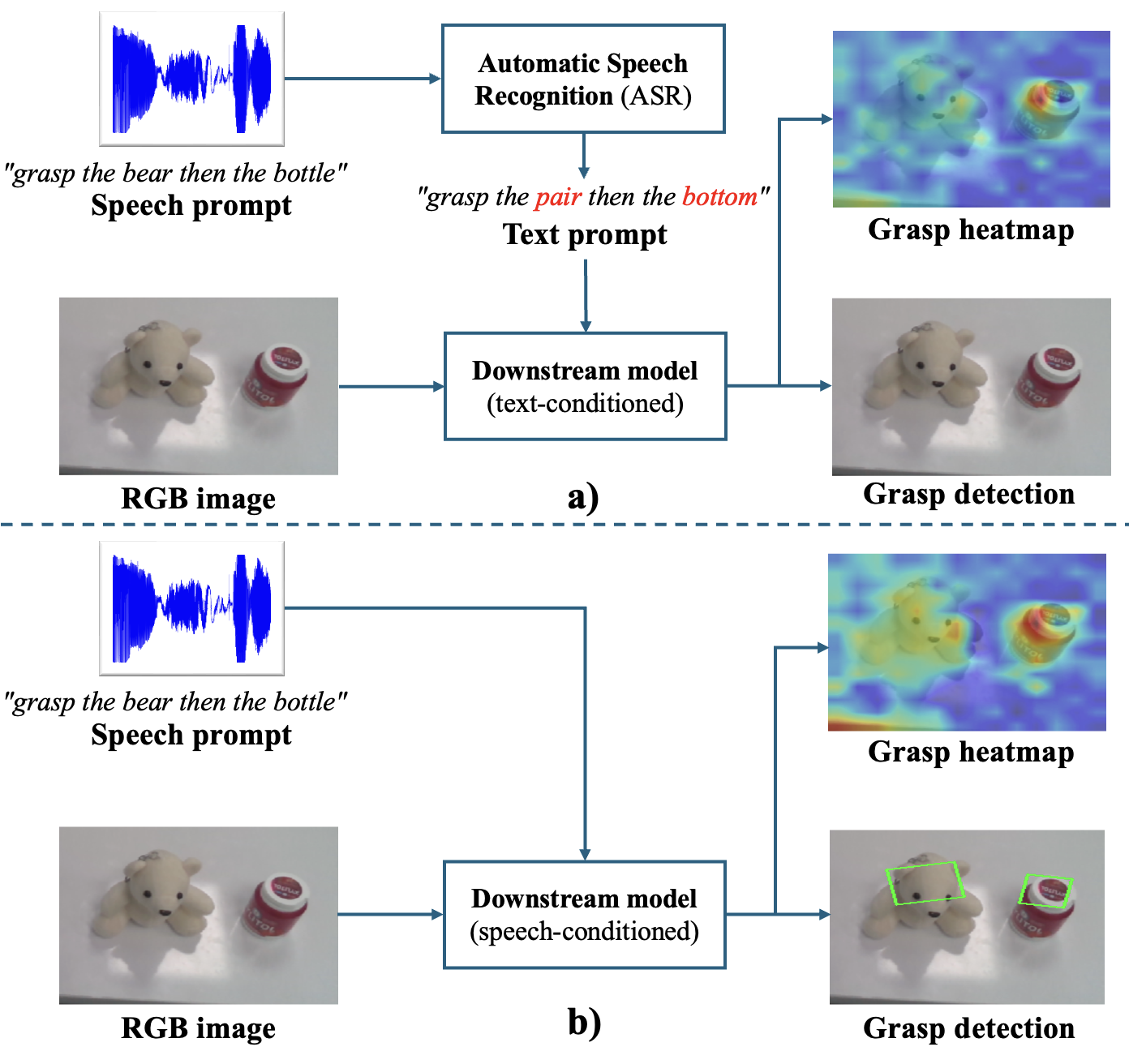}
    \caption{Methods for transferring text-conditioned models to speech. a) uses an Automatic Speech Recognition (ASR) model to transcribe the speech, but it mistranscribes, causing complete failure of the downstream task. b) is trained directly on speech, bypassing the discrete mistranscription problem.}
    \label{fig_intro}
\end{figure}

Humanoid robots increasingly require multi-modal understanding for natural interaction with humans \cite{HumanIntro}. To support such capability, vision-language models (VLMs) have emerged as the predominant foundation for language-conditioned perception. However, these models generally assume textual language inputs \cite{HumanText1}, despite speech being a more natural and intuitive communication modality. While foundation image-speech models \cite{SpeechFound1, SpeechFound2, SpeechFound3, SpeechFound4} have demonstrated robust joint representations, adapting them to downstream robotic tasks may require collecting large-scale speech datasets together with corresponding task annotations, followed by non-trivial architectural redesigns and computationally expensive fine-tuning.

A practical alternative is to endow existing image-text models, which are already well established for the downstream task, with speech capability. The simplest solution is to employ a pre-trained Automatic Speech Recognition (ASR) model to convert speech into text for the text-conditioned downstream model. However, such cascaded pipelines are prone to higher latency and error propagation \cite{Connector}. As illustrated in Figure \ref{fig_intro}a, where grasp detection is the downstream task, if the ASR mistranscribes the prompt, the grasp attempt will fail entirely as the target object is not present. In contrast, models trained directly on speech prompts may offer improved robustness to speaker and acoustic variations \cite{SpeechSurvey}. However, as mentioned, fine-tuning such models often incurs prohibitive dataset and computation requirements.

The above discussion motivates our work: transferring a well-established text-conditioned robotic system to operate directly on speech without using large speech datasets. In this paper, we choose ALBEF \cite{ALBEF} as the pre-trained VLM and language-driven grasp detection \cite{Binpicking1, GraspAnything} as the downstream task, as a case study. Firstly, we investigate the representation gap introduced in the joint ALBEF embedding space when replacing text with speech. Our central finding is that a lightweight MLP-based projector is sufficient to adapt ALBEF to the new speech modality, without requiring large-scale data. Additionally, we perform diagnostic analyses showing that the adaptation retains semantic discrimination and robustness to unseen acoustics and speaker variations. Motivated by these findings, we introduce Speech2Grasp, a framework for data-efficient transfer of text-conditioned grasp detection to speech. Our simulation and real-world experiments on a humanoid robot demonstrate that Speech2Grasp achieves equal or better grasp detection rates than cascaded ASR-based pipeline, while incurring less latency due to bypassing the transcription step. Furthermore,  Speech2Grasp only requires small-scale speech data for fine-tuning. More broadly, our findings suggest that lightweight modality transfer can provide a practical paradigm for extending established text-conditioned systems to speech. In summary, our contributions are as follows: 
\begin{itemize}
    \item We conduct diagnostic analyses showing that ALBEF can be adapted to speech using robust, data-efficient adaptation based on an MLP-based projector.
    \item We introduce Speech2Grasp, a framework for data-efficient transfer of text-conditioned grasp detection to speech. Real-world humanoid experiments show that Speech2Grasp outperforms cascaded ASR-based pipeline, while incurring less latency.
\end{itemize} 

\section{Related Works}

\textbf{Cross-Modal Transfer to Speech.} Generally, these works distill knowledge from large-scale pre-trained text models. Ni et al. \cite{AdaptiveKD} use an attention-based strategy to align text-speech representations for spoken language understanding, using only 10 hours of audio. Xie et al. \cite{AFD-SLU} show that such alignment can be performed using a lightweight MLP-based network and an adaptive loss between the text and speech embeddings, using only 4K speech samples. These works establish that text and speech share common semantic representations, enabling data-efficient transfer between these modalities. However, such transfer has been relatively under-explored for robotic perception.

\textbf{Efficient Adaptation for Grasp Detection.} Generally, these works transfer representations from a large teacher network to a compact student. KufeNet \cite{unequal} distills large-scale knowledge into a lightweight model with unequal RGB and depth contributions. LiteGrasp \cite{LiteGrasp} uses feature-level distillation with pseudo-label supervision from the teacher model. PDCNet \cite{PDCNet} integrates partial convolutions and the wavelet transform to achieve feature-aware compact representations. Although these works enable lightweight deployment without performance degradation, extensions to modality transfer remain largely under-explored.

\textbf{Language-conditioned Grasp Detection.} Vuong et al. \cite{GraspAnything} introduce Grasp-Anything, a large-scale grasp detection dataset with natural-language text prompts. Grasp-Anything++ \cite{GraspAnythingpp} extends with part-level annotations. GraspSAM \cite{GraspSAM} unifies prompt-based segmentation and grasp detection by adapting a pre-trained SAM \cite{SAM}. Grasp-Anything-6D \cite{GraspAnything-6DoF} extends 2D datasets with annotated 3D point clouds. However, these methods assume text as language input, leaving direct speech-guided grasp detection largely unexplored.

\section{Diagnostic Analyses of ALBEF Text–Speech Representations}

\subsection{The ALBEF module}

\begin{figure}[t]
    \centering
    \includegraphics[width=0.8\linewidth]{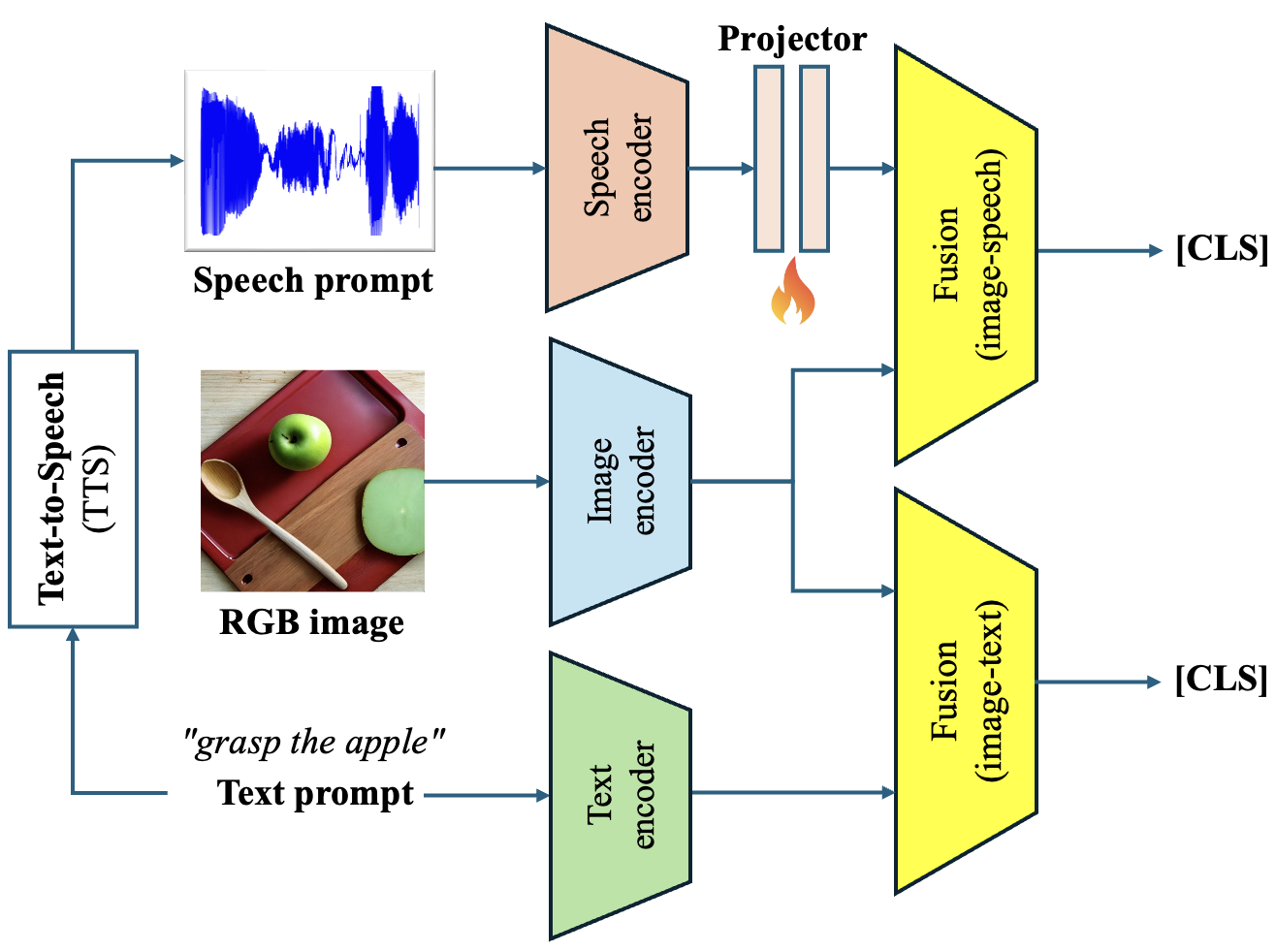}
    \caption{ALBEF \cite{ALBEF} fusion between image, text and speech modalities. The bottom two blocks show the frozen, teacher modules that perform image-text fusion. Our diagnostic study analyzes the representation gap introduced when replacing text with the corresponding speech, as in the top two blocks.}
    \label{fig_albef}
\end{figure}

\textbf{AL}ign \textbf{BE}fore \textbf{F}use (ALBEF) is a pre-trained vision-language model designed to learn semantically aligned image-text representations \cite{ALBEF}. As illustrated in the bottom two blocks of Figure~\ref{fig_albef}, the ALBEF fusion module combines image and text features to produce the \texttt{[CLS]} embeddings that serve as their joint representation. Because the image and text modalities produce heterogeneous feature distributions, ALBEF employed an alignment objective during pre-training to project both modalities into a shared latent space before fusion, hence its name. Motivated by the need to replace text with the more natural and intuitive speech modality for human-robot interaction, this Section investigates whether speech-conditioned representations can remain aligned with this shared latent space.

\subsection{Text-Speech Representation Gap in the ALBEF Space}

\begin{figure}[t]
    \centering
    \includegraphics[width=0.9\linewidth]{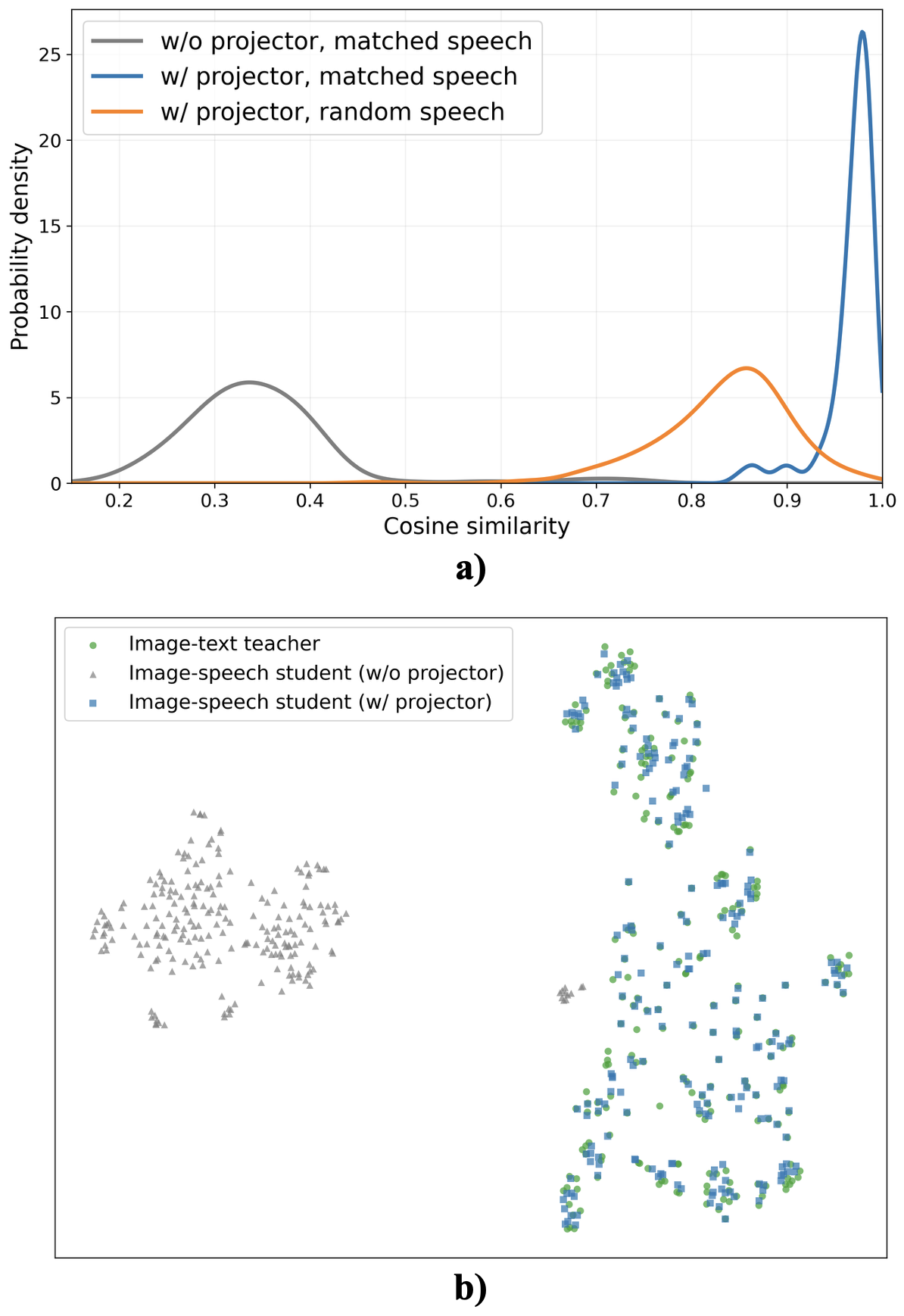}
    \caption{Effects of the trained projector on text-speech representation gap. a) The projector enables the image-speech student to closely align with the image-text teacher while retaining semantic discrimination. b) t-SNE visualizations show the projected student embeddings closely overlap the teacher, while the unprojected embeddings remain separated.}
    \label{fig_diag_secB}
\end{figure}

Starting from the pre-trained ALBEF, we replace the text modality with speech, while keeping the image input unchanged. The image-text pairs were obtained from Grasp-Anything++ \cite{GraspAnythingpp}, a language-conditioned grasp detection dataset that provides RGB images paired with textual grasp instructions. We use Kitten-TTS \cite{KittenTTS} to obtain the corresponding speech instructions. In total, we obtain 15K aligned image-text-speech samples, with the train-validation-test split 80\%-20\%-20\%. Afterwards, as illustrated in the top two blocks of Figure \ref{fig_albef}, a pre-trained Whisper \cite{Whisper} encoder extracts the speech embeddings, which are subsequently passed through a trainable lightweight MLP-based projector. The projector maps the speech embeddings into the shared ALBEF latent space, and is trained through the following objective based on cosine similarity (CS):

\begin{equation} \label{loss_KD}
\mathcal{L}_{\text{KD}}
= 1 -
\frac{\mathbf{e}_{\text{S}}^\top \mathbf{e}_{\text{T}}}
{\|\mathbf{e}_{\text{S}}\|_2 \, \|\mathbf{e}_{\text{T}}\|_2}
\end{equation}
where $\mathbf{e}_{\text{S}}$ and $\mathbf{e}_{\text{T}}$ denote the student (image-speech) and teacher (image-text) \texttt{[CLS]} embeddings, respectively. The ``student'' and ``teacher'' terminology reflects the fact that the image-speech branch is trained to reproduce the joint representation generated by the frozen image-text branch. Therefore, this objective is considered a knowledge distillation (KD) loss based on representation alignment \cite{AFD-SLU}.

Figure \ref{fig_diag_secB}a shows the kernel density estimate (KDE) of the CS between the student and teacher's \texttt{[CLS]} embeddings. Three variants of the student are evaluated: (i) without the projector using matched speech (gray curve), (ii) with the projector using matched speech (blue), and (iii) with the projector using randomly paired speech (orange). As illustrated by the gray curve, directly using the speech embeddings without the projector results in a substantial representation gap. The average CS is $0.346 \pm 0.088$, indicating a weak alignment. This result is intuitive, because the ALBEF fusion module was pre-trained on large-scale image-text pairs, rather than image-speech. However, the CS remains positive, most likely because the image modality is unchanged. The blue curve indicates that the trained projector successfully maps the speech embeddings into the teacher's latent space. The average CS is $0.968 \pm 0.027$, indicating a very strong alignment using only a lightweight MLP-based projector. We hypothesize that this strong alignment arises from three complementary factors: i) the shared image input provides a common visual context for both the teacher and student, ii) the pre-trained speech and text encoders already capture similar high-level semantics, leaving the projector to learn primarily the modality-specific distribution mismatch, and iii) alignment is evaluated using the fused \texttt{[CLS]} embeddings, which summarize global information and are therefore easier to align than token-level representations. Interestingly, comparing the blue and orange curves, where the latter corresponds to randomly paired speech, shows that the projector preserves semantic discrimination rather than collapsing different speech inputs to similar embeddings. The average CS of random pairs is $0.835 \pm 0.069$, which remains considerably lower than that of matched pairs. Furthermore, the matched model yields a higher CS than the shuffled model in 98.2\% of the test instances. Before projection, the former only beats the latter in 50.7\% of the test instances. Finally, the broad distribution of the orange curve highlights much greater variability while the blue curve remains highly concentrated, indicating that the projector does not produce uniformly similar embeddings for mismatched instructions.

Figure \ref{fig_diag_secB}b presents the t-SNE \cite{t-SNE} visualizations of the \texttt{[CLS]} embeddings for i) the teacher (green), ii) the unprojected (gray) and iii) the projected student (blue). The projected student embeddings closely overlap the teacher embeddings, whereas the unprojected embeddings remain separated. These results indicate that the projector aligns the student embedding with the teacher structurally, complementing the high pairwise CS.

\subsection{Generalization to Unseen Speakers and Acoustic Conditions} \label{sec_prj_robust}

\begin{figure}[t]
    \centering
    \includegraphics[width=0.9\linewidth]{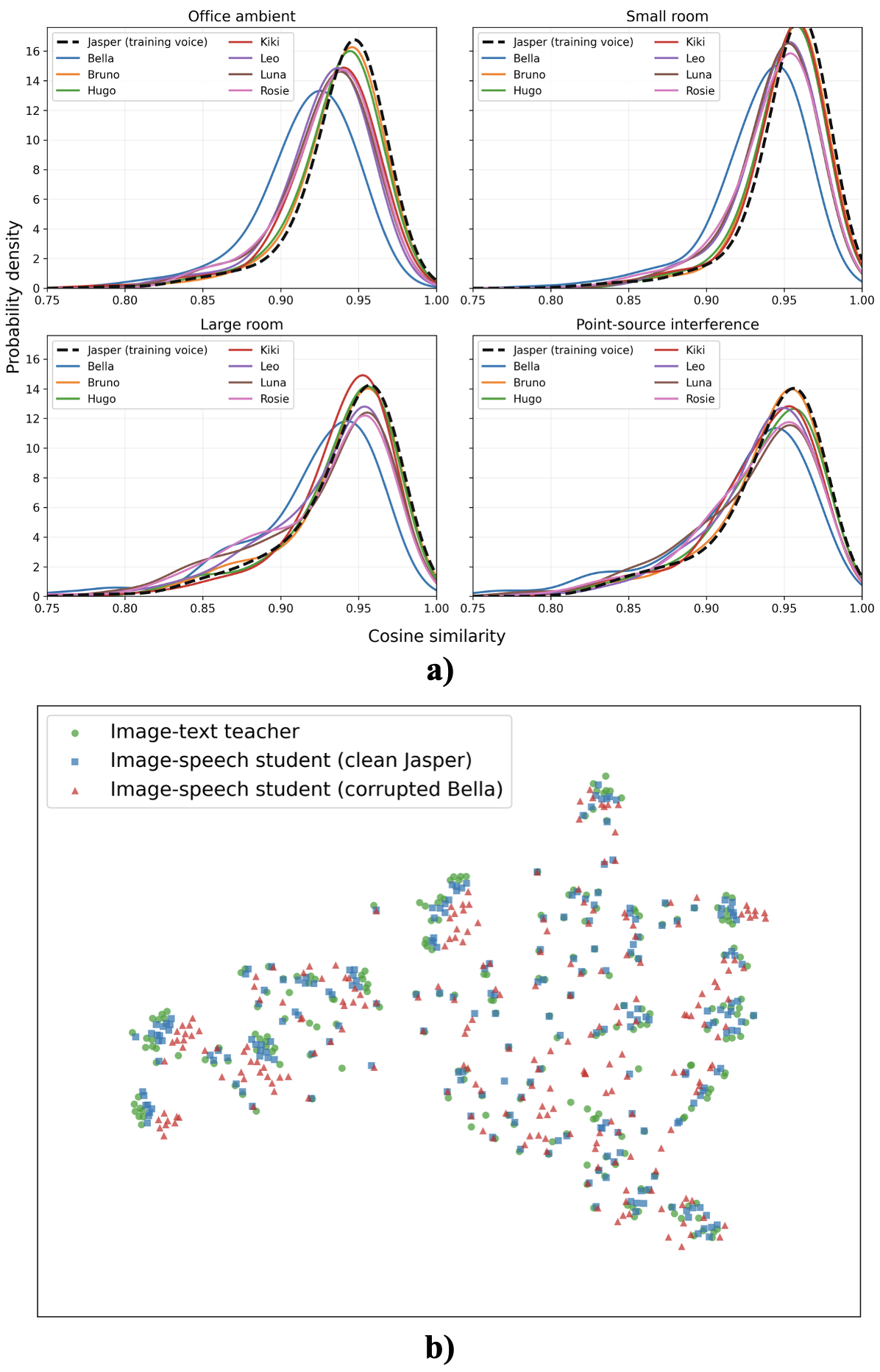}
    \caption{Generalization of the trained projector to real-world conditions. a) The projector enables the student to closely align with the teacher, despite large acoustics and speaker variations. b) t-SNE visualizations show both corrupted and clean student embeddings closely overlap the teacher.}
    \label{fig_diag_secC}
\end{figure}

The previous analyses demonstrate that the image-speech and image-text \texttt{[CLS]} embeddings are sufficiently aligned that a lightweight MLP-based projector can effectively bridge the remaining representation gap while retaining semantic discrimination. However, real-world speech is often affected by many factors, among which are i) acoustic conditions, such as environmental noise and room reverberation, and ii) speaker variability. Therefore, in this Section, we investigate whether the projector, which was trained exclusively on clean speeches of a single voice, generalizes to these conditions.

\begin{figure*}[t]
    \centering
    \includegraphics[width=0.90\linewidth]{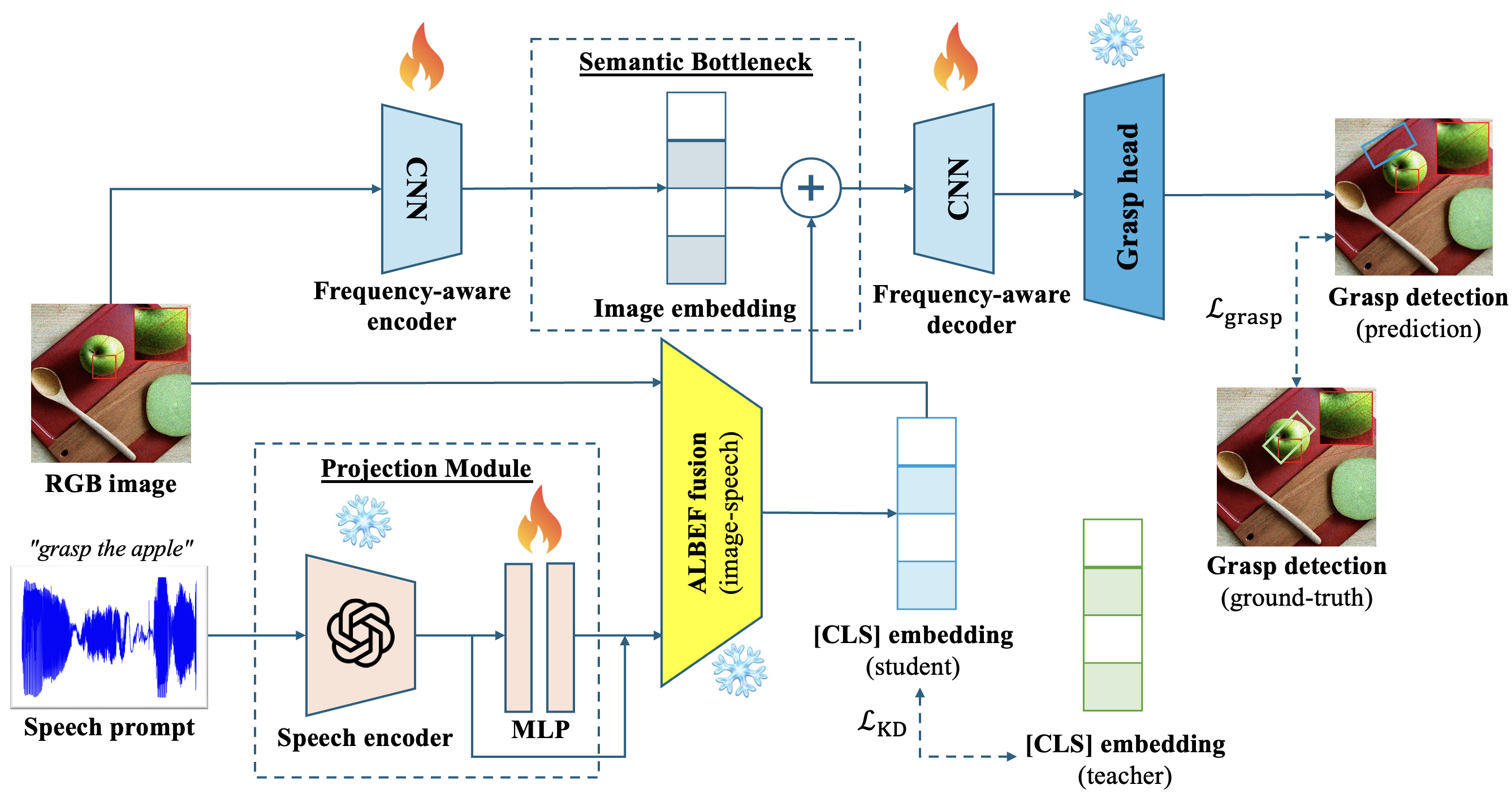}
    \caption{Overview of the Speech2Grasp framework for data-efficient transfer of text-conditioned grasp detection to speech. An MLP-based projector maps speech embeddings into the ALBEF \cite{ALBEF} latent space. A knowledge distillation loss, $\mathcal{L}_\text{KD}$, encourages the student's projected \texttt{[CLS]} image-speech embeddings to match the teacher's image-text embeddings, enabling ALBEF's reuse without large-scale paired speech annotations. The resulting \texttt{[CLS]} embedding is injected into the visual semantic bottleneck to guide grasp prediction, supervised by the task loss $\mathcal{L}_\text{grasp}$.}
    \label{fig_arch}
\end{figure*}

\begin{figure}[thbp]
    \centering
    \includegraphics[width=0.9\linewidth]{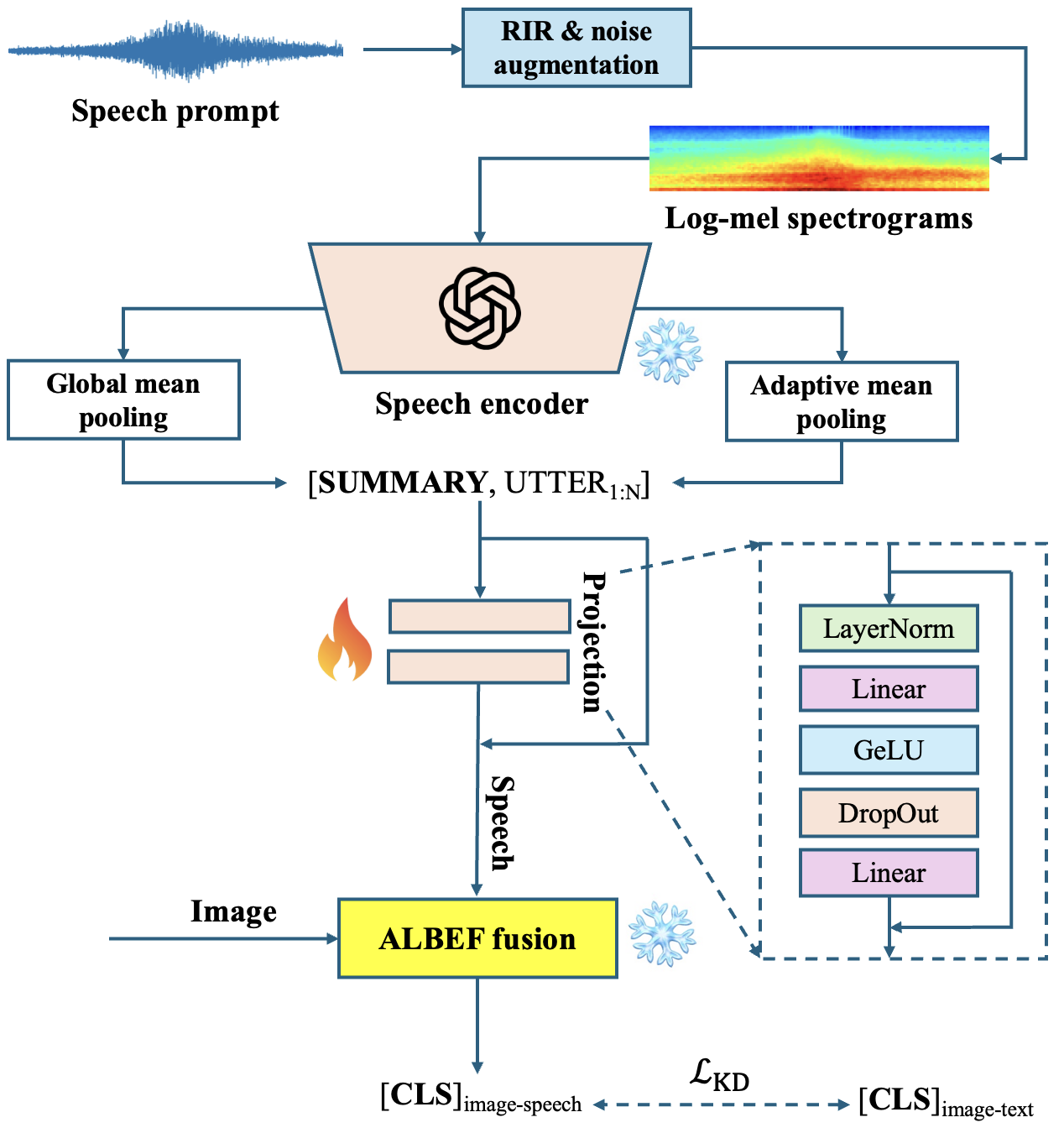}
    \caption{Speech encoding and projection. The augmented speech is converted into log-mel spectrograms before being encoded by Whisper [4]. Global or adaptive mean pooling is then applied to produce a token sequence similar to the text-based BERT [8] representation.}
    \label{fig_speech_encoding}
\end{figure}

\begin{figure}[thbp]
    \centering
    \includegraphics[width=0.9\linewidth]{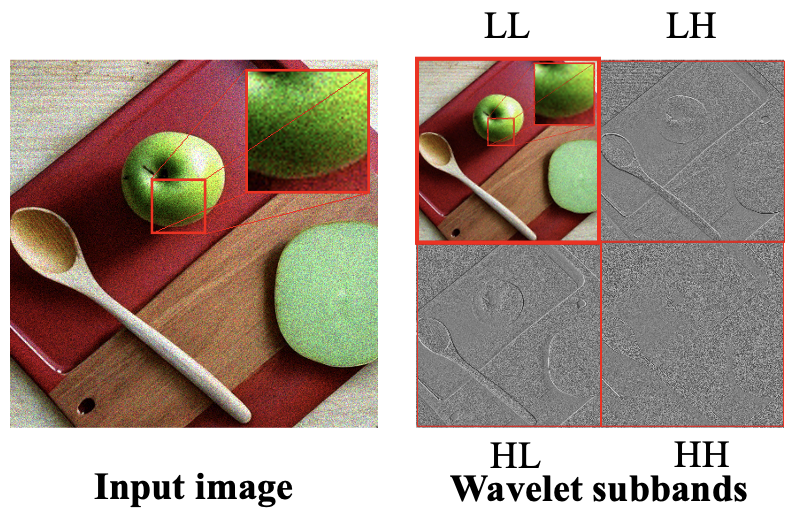}
    \caption{The DWT decomposes a noisy input image into four wavelet subbands (LL, LH, HL, HH). They capture coarse, horizontal, vertical and diagonal information, respectively. The LL is less noisy, the LH and HL capture object boundaries and the HH subband is dominated by noise.}
    \label{fig_dwt}
\end{figure}

Firstly, from the 3K aligned image-text pairs in the test set, we additionally synthesize 21K speech samples by rendering each text instruction using the seven unseen voices provided by Kitten-TTS. Afterwards, each sample is acoustically perturbed according to the following model:
\begin{equation} \label{eq_model}
    y(t) = x(t) * h(t) + \alpha n(t),
\end{equation}
where $x(t)$ denotes the clean speech, $*$ denotes convolution and $h(t)$ is the room impulse response (RIR). $n(t)$ represents the environmental noise, and $\alpha$ is a scaling factor for a signal-to-noise ratio of 8\,dB, where the speech remains intelligible while noticeably degraded. We construct four conditions using combinations of RIRs and noise samples obtained from \cite{OpenSLR}. In total, we obtain 84K unseen speech samples that vary in both acoustics and speaker, together with the original 15K.

The subplots in Figure \ref{fig_diag_secC}a present the KDE curves for these conditions. Each KDE curve corresponds to a speaker, where the dotted black one denotes the training voice (``Jasper''). Overall, these results demonstrate graceful degradation: although real-world speech reduces the CS, the projector preserves a high degree of alignment, as the curves only move slightly to the left. For the worst-case scenario (``Bella'' voice, point-source interference), the average CS is $0.915 \pm 0.070$, which still indicates a good alignment. The underlying explanation is identical to the previous Section: shared image inputs, shared semantic representations, and the ALBEF \texttt{[CLS]} embedding. Whisper's large-scale pre-training on acoustically diverse speech may also contribute to this robustness. Figure \ref{fig_diag_secC}b presents the t-SNE visualizations for i) the teacher (green), ii) the student (blue), and iii) the student evaluated in the worst-case scenario (red). While the worst-case embeddings exhibit moderate shifts, they remain structurally similar to both the teacher and the clean student embeddings.

\subsection{Implications of the Diagnostic Analyses}

Collectively, the analyses indicate that a lightweight MLP-based projector can effectively align the ALBEF \texttt{[CLS]} embeddings of the image-text teacher and image-speech student, while preserving semantic discrimination and robustness under the evaluated acoustic perturbations. This suggests that the semantic knowledge of ALBEF-based, text-conditioned models can be transferred to speech through lightweight adaptation. Furthermore, successful generalization from 9K paired image-speech training samples to 84K unseen and corrupted speech samples suggests that such adaptation may not require prohibitively large speech datasets. Finally, such systems might operate directly from speech, without the need for transcription, while remaining robust to real-world conditions. These observations motivate the architectural choices and dataset design presented in the next Section for data-efficient transfer of text-conditioned grasp detection to speech.

\section{Methodology}

\subsection{Problem Statement}

Building upon the previous analyses, we formulate the problem of data-efficient transfer of ALBEF-based text-conditioned grasp detection to speech. Let $\mathbf{I}$ denote an RGB image, $T$ a textual grasp instruction, and $S$ its corresponding speech instruction. Given a pre-trained grasp detector $f(\mathbf{I},T) \rightarrow \mathbf{G}$, where $\mathbf{G}$ denotes the grasp map prediction, our objective is to construct a speech-conditioned counterpart $g(\mathbf{I},S) \rightarrow \mathbf{G}$
that operates directly on speech. The adaptation should preserve the teacher's grasp capability while requiring only a modest amount of paired image-speech data.

\subsection{Overall Framework}

Figure~\ref{fig_arch} presents an overview of the proposed Speech2Grasp framework. We first initialize the student model $g(\mathbf{I},S)$ from the pre-trained text-conditioned grasp detector $f(\mathbf{I},T)$, namely LGD \cite{GraspAnythingpp}. As illustrated in Figure \ref{fig_arch}, Speech2Grasp consists of a visual branch (upper) and a speech branch (lower). To inherit grasp capability from the teacher, in the visual branch, the grasp head is kept frozen, while the U-shaped encoder-decoder is only lightly fine-tuned. To adapt to speech, the speech branch consists of a pre-trained Whisper \cite{Whisper} encoder, as well as an MLP-based projector to map the speech embeddings into the ALBEF latent space. As demonstrated in the analyses, the high pairwise CS means the projected speech embeddings behave analogously to text embeddings. Therefore, similar to the teacher, they can be injected into the same semantic bottleneck for vision-language fusion, using the original fusion strategy. Finally, the speech-aware, fused embeddings are passed through the decoder and grasp head for final grasp map prediction. The total training loss is as follows:
\begin{equation}
\mathcal{L} =
\mathcal{L}_{\text{grasp}} + \lambda_{\text{KD}}\mathcal{L}_{\text{KD}}
\end{equation}
where $\mathcal{L}_{\text{grasp}}$ is the task grasp loss and $\mathcal{L}_{\text{KD}}$ is the knowledge distillation loss introduced in Equation \eqref{loss_KD}. $\mathcal{L}_{\text{grasp}}$ is detailed as follows:
\begin{equation}
\mathcal{L}_{\text{grasp}}
=
\frac{1}{HW}
\sum_{i=1}^{H}\sum_{j=1}^{W}
\left\|
\mathbf{G}_{ij}^{\text{pred}} -
\mathbf{G}_{ij}^{\text{gt}}
\right\|_2^2
\end{equation}
$H$ and $W$ denote the image height and width, and 
$\mathbf{G}^{\text{pred}}$ and $\mathbf{G}^{\text{gt}}$ represent the predicted 
and ground-truth grasp maps, respectively. $\lambda_{\text{KD}}$ is the balancing term. We employ an off-ramp linear scheduler for $\lambda_{\text{KD}}$ so that training initially emphasizes representation alignment, and gradually shifts to focus solely on the task loss. This strategy allows the speech embeddings to rapidly adapt to the ALBEF latent space, while emphasizing the grasp supervision later.

\subsection{Speech Encoding and Projection}

Figure \ref{fig_speech_encoding} provides details about the speech encoding and projection. Firstly, the speech inputs are augmented according to Equation \eqref{eq_model}, although we sample all possible RIRs and noise samples. The augmentation is performed on the fly during training to simulate diverse acoustic conditions without increasing dataset storage. Afterwards, the augmented speech signals are converted into log-mel spectrograms \cite{SpeechBook}, the expected inputs for Whisper \cite{Whisper}. We then convert the Whisper outputs into a BERT-like \cite{BERT} token sequence, where the \texttt{[SUMMARY]} token is obtained via global mean pooling and captures the global speech representation. The remaining \texttt{[UTTER]} tokens are obtained via adaptive pooling to preserve localized utterance-level information. The trainable projector allows the pseudo-text sequence to replace the actual one expected by ALBEF fusion. The architecture of the projector is detailed in Figure \ref{fig_speech_encoding}, where the skip connection is used to reduce overfitting. 

\subsection{Frequency-aware Encoder and Decoder} \label{sec_freq_aware}

To maximize Speech2Grasp's deployability, we improve its robustness against camera noise often encountered in practical scenarios. We hypothesize that the encoder-decoder in the visual branch should be frequency-aware. Precisely, it should emphasize high-frequency (HF) components corresponding to object boundaries, which are essential for accurate grasp localization. Furthermore, it should attenuate HF noise arising from camera sensors.

Following previous works \cite{DWTNeRF, DWTGS, AutoOpti3DGS, WaveletGaussian, EUSIPCO}, we use the Discrete Wavelet Transform (DWT) to perform such frequency-aware decomposition. As illustrated in Figure \ref{fig_dwt}, the DWT decomposes an input image into four wavelet subbands (LL, LH, HL, HH), which capture coarse, horizontal, vertical and diagonal information, respectively. Instead of directly using the RGB image $\mathbf{I}\in\mathbb{R}^{H\times W\times 3}$, our framework consumes an input of dimension $\mathbb{R}^{H/2\times W/2\times 9}$, which is obtained by concatenating the LL, LH, and HL subbands while discarding the noise-dominant HH. The explicit preservation of the LH and HL subbands retains the essential object boundaries, while discarding the HH suppresses noise without significantly sacrificing boundary information. Since the DWT performs downsampling, each subband is half-resolution. Therefore, although the input channel dimension increases, the computation overhead remains modest, following previous works \cite{WaveDiff}. While the encoder-decoder remains trainable, only its first few layers are optimized to adapt to the wavelet subbands, thus preserving grasp capability.

\section{Experiments}

\subsection{Simulation Experiments} \label{sec_exp_sim}

We evaluate Speech2Grasp on the Grasp-Anything++ \cite{GraspAnythingpp} dataset. The teacher model, LGD, is provided in the corresponding GitHub repository \cite{GraspAnythingpp}. We use KittenTTS \cite{KittenTTS} to synthesize speech prompts according to the object frequency in the LVIS \cite{LVIS} taxonomy. Among these objects, 70\% correspond to ``Seen'' (S) category and 30\% to ``Unseen'' (U) category. Our TTS generation produces approximately 15K speech prompts, totaling about 14 hours of audio. To evaluate retention of grasp performance, we use the success rate across S and U categories, following \cite{GraspAnythingpp}. We also use the harmonic mean (H) to summarize the overall success rate \cite{Harmonic}. In addition, we compare the inference latency required to produce the ALBEF \texttt{[CLS]} embedding, in miliseconds and for the same speech. The number of training samples is also recorded.

Table \ref{tab_results} presents the results, which were averaged across 10 re-runs. The first row corresponds to a cascaded pipeline where the clean speech is first transcribed using LiteASR \cite{LiteASR}, an efficient ASR model that achieves performance competitive with large models. This pipeline is the natural alternative for speech adaptation. The second row is our framework and the bottom row is the text-conditioned teacher. The cascaded pipeline suffers from slight performance drops even for clean speech, due to the occasional mistranscriptions (examples provided in Table \ref{tab_asr_errors}). Our Speech2Grasp achieves similar performance to the text-only teacher, indicating successful transfer. However, because the transcription step is eliminated, the latency is reduced threefold. Finally, due to the shared semantic information, the adaptation only requires 15K samples. As indicated in Table \ref{tab_kd_results}, the final CS is 0.94, indicating that the projector effectively maps speech inputs to text-like embeddings expected by the teacher.

\begin{table}[thbp]
\centering
\begin{tabularx}{\linewidth}{l*{5}{>{\centering\arraybackslash}X}}
\toprule
Method & S($\uparrow)$ & U$(\uparrow)$ & H$(\uparrow)$ & Latency $(\downarrow)$ & \#S$(\downarrow)$ \\
\midrule
LiteASR $\rightarrow$ LGD & 0.34 & 0.33 & 0.34 & 102.2 & 1M \\

Speech2Grasp (Ours)
& \textbf{0.36} & \textbf{0.33} & \textbf{0.35} & \textbf{36.6} & \textbf{15K} \\
\midrule
LGD & \textbf{0.36} & \textbf{0.33} & \textbf{0.35} & -- & 1M \\
\bottomrule
\end{tabularx}
\caption{Results of Data-efficient Transfer of Grasp Detection}
\label{tab_results}
\end{table}

We then measure robustness against simulated camera noise. We corrupt the input images in the test set using a Poisson-Gaussian noise model \cite{PoissonGaussian}. We consider two noise regimes. The \textit{``mild''} setting uses $\lambda = 30, \sigma = 0.01$, while the \textit{``severe''} setting uses $\lambda = 10, \sigma = 0.02$. Table \ref{tab_results_noise} presents the results across 10 re-runs. Generally, without the DWT, the grasp detection performance degrades quickly, even when the speech prompts are clean. This occurs because of the fused architecture, where noisy visual features can propagate through the fusion module and degrade the final grasp predictions. Our DWT-integrated Speech2Grasp is more robust because i) the LL is less noisy, and ii) the HH is discarded, although the model was not trained on noisy images. In the case of \textit{``mild''} noise, our Speech2Grasp performs similar to the noiseless case in Table \ref{tab_results}, as the LH and HL preservation retains the object boundaries crucial for grasp detection, while the overall geometry is still captured by the LL.

\begin{table}[htbp]
\centering
\begin{tabularx}{\linewidth}{l l *{3}{>{\centering\arraybackslash}X}}
\toprule
Mode & Method & S($\uparrow$) & U($\uparrow$) & H($\uparrow$)\\
\midrule

\multirow{3}{*}{Mild}
& LiteASR $\rightarrow$ LGD (w/o DWT) & 0.31 & 0.31 & 0.31 \\
& Speech2Grasp (Ours - w/ DWT) & \textbf{0.34} & \textbf{0.33} & \textbf{0.34} \\
& LGD (w/o DWT) & 0.32 & 0.31 & 0.32 \\

\midrule

\multirow{3}{*}{Severe}
& LiteASR $\rightarrow$ LGD (w/o DWT) & 0.25 & 0.24 & 0.24 \\
& Speech2Grasp (Ours - w/ DWT) & \textbf{0.31} & \textbf{0.27} & \textbf{0.29} \\
& LGD (w/o DWT) & 0.27 & 0.24 & 0.25 \\

\bottomrule
\end{tabularx}
\caption{Results under simulated camera noise}
\label{tab_results_noise}
\end{table}

Finally, in Table \ref{tab_kd_results}, we investigate the effects of the number of speech samples on representation alignment, using 7.5K, 11K and 15K speech samples. Generally, increasing the number of speech samples improves representation alignment, and thus the grasp detection performance. 


\begin{table}[htbp]
\centering
\begin{tabularx}{\linewidth}{l*{4}{>{\centering\arraybackslash}X}}
\toprule
Method & S$(\uparrow)$& U$(\uparrow)$& H$(\uparrow)$& CS$(\uparrow)$\\
\midrule
Speech2Grasp (7.5K) & 0.31 & 0.29 & 0.30 & 0.83 \\
Speech2Grasp (11K) & 0.35 & 0.30 & 0.32 & 0.87 \\
Speech2Grasp (15K) & \textbf{0.36} & \textbf{0.33} & \textbf{0.35} & \textbf{0.94} \\
\midrule
LGD \cite{GraspAnythingpp} & \textbf{0.36} & \textbf{0.33} & \textbf{0.35} & -- \\
\bottomrule
\end{tabularx}
\caption{Results under Varying Number of Speech Samples}
\label{tab_kd_results}
\end{table}

\begin{table}[t]
\centering
\scalebox{0.95}{%
\begin{minipage}{\columnwidth}
\begin{tabularx}{\columnwidth}{XX}
\toprule
Speech prompt & ASR mistranscription \\
\midrule

\textit{``Pick up spoon by its bowl shape''}
& \textit{``Pick up spoon \textcolor{red}{bites bowl shaped}''} \\
\textit{``Grasp plant at its leaves''}
& \textit{``\textcolor{red}{Grass} plant at its leaves''} \\
\textit{``Grasp scissors at its handles''}
& \textit{``Grasp \textcolor{red}{sisters} at its handles''} \\

\midrule

\textit{``Grasp the bear then the bottle''}
& \textit{``Grasp the \textcolor{red}{pair} then the bottle''} \\
& \textit{``Grasp the bear then the \textcolor{red}{bottom}''} \\

\bottomrule
\end{tabularx}
\end{minipage}%
}
\caption{Examples of ASR mistranscriptions of clean speech (top) and real-world speech (bottom)}
\label{tab_asr_errors}
\end{table}

\begin{table}[htbp]
\centering
\begin{tabularx}{\linewidth}{l l >{\centering\arraybackslash}X >{\centering\arraybackslash}X}
\toprule
Person & Method & Single-object($\uparrow$) & Multi-object($\uparrow$)\\
\midrule
\multirow{2}{*}{\#1}
& LiteASR $\rightarrow$ LGD & 0.59 $\pm$ 0.03 & 0.54 $\pm$ 0.06 \\
& Speech2Grasp (Ours) & \textbf{0.70} $\pm$ 0.02 & \textbf{0.61} $\pm$ 0.05 \\
\midrule
\multirow{2}{*}{\#2}
& LiteASR $\rightarrow$ LGD & 0.61 $\pm$ 0.04 & 0.56 $\pm$ 0.07 \\
& Speech2Grasp (Ours) & \textbf{0.72} $\pm$ 0.03 & \textbf{0.63} $\pm$ 0.06 \\
\bottomrule
\end{tabularx}
\caption{Real-world Grasp Detection Results on a Humanoid Robot}
\label{tab_real_results}
\end{table}

\begin{figure}[thbp]
    \centering \includegraphics[width=0.75\linewidth]{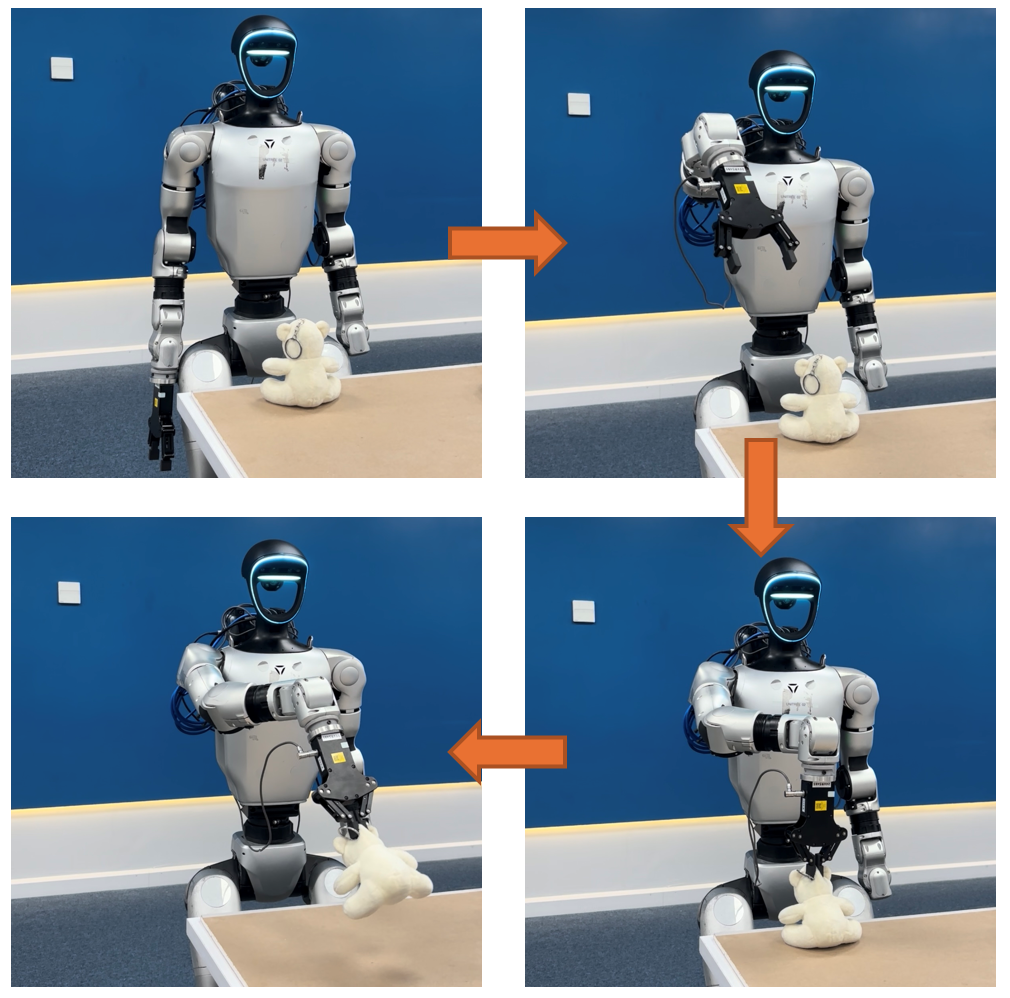}
    \caption{Grasp sequence of the humanoid robot using Speech2Grasp.}
    \label{fig_robot_op}
\end{figure}

\subsection{Real-world Robotic Experiments} Our real-world robotic experiments, which were conducted on a Unitree humanoid robot \cite{UnitreeWebsite}, demonstrate the feasibility of Speech2Grasp for practical deployment. Our speech commands were recorded in an uncontrolled office environment using a standard laptop microphone. We evaluated two scenarios: single-object and multi-object. In the latter, the robot is instructed to grasp multiple unseen objects sequentially (e.g., \textit{``grasp the bear then the bottle''}). Specifically, the robot executes the grasp with the highest confidence, and inference is then repeated until all objects have been grasped. None of the objects are seen during training or validation.

We repeated the grasp attempts 30 times for each case, and for two non-native English speakers. The total attempts were 120. We present the results in Table \ref{tab_real_results}. The cascaded pipeline requires a transcription step. However, because the real-world speech commands are short and lack contextual cues, they are more prone to mistranscriptions than clean speech. Table \ref{tab_asr_errors} provides some examples. Once such errors occur, the grasp attempt will fail completely because the target object is not present, as illustrated in Figure \ref{fig_intro}a and the corresponding video submission. Our Speech2Grasp, natively trained on speech, provides a better grasp detection rate due to its tolerance to mild acoustic and speaker variations. Note that the improvement in the simulation experiments at Section \ref{sec_exp_sim} is modest because the speech prompts are clean and synthesized, allowing ASR to perform well. The larger gains in the real-world experiments arise because the ASR produces more mistranscriptions. Speech2Grasp's robustness can be attributed to the speech projector, whose resilience under much more challenging acoustic conditions was demonstrated in the diagnostic study at Section \ref{sec_prj_robust}.

\section{Conclusion}

In this paper, we first conduct diagnostic analyses to show that a data-efficient, lightweight MLP-based projector effectively aligns speech embeddings into the ALBEF latent space, while preserving semantic discrimination and robustness to acoustic and speaker variations. Motivated by these findings, we propose Speech2Grasp, a data-efficient framework that transfers text-conditioned grasp detection to speech. Simulation experiments show that Speech2Grasp achieves grasp performance comparable to the text-based teacher, while being data-efficient and faster in inference compared to an ASR-based pipeline. Real-world experiments on a humanoid robot demonstrate its deployability under real-world acoustic conditions, while avoiding discrete mistranscription errors often occurring in real-world speech.

\textbf{Discussion and Future Works}. Firstly, it would be valuable to investigate whether the proposed speech transfer paradigm generalizes to vision-language architectures with multi-modal fusion strategies different from that of ALBEF. Secondly, the teacher-student framework incurs substantial GPU memory overhead due to the forward passes of both teacher and student networks during training. Future work will investigate cached knowledge distillation by pre-computing teacher embeddings, enabling more memory-efficient speech transfer. Broadly speaking, we hope our findings suggest a practical paradigm for extending established text-conditioned systems to speech, supporting the growing demand for natural speech interaction in humanoid robots.

\bibliographystyle{IEEEtran}
\bibliography{main}

@misc{DWTNeRF,
      title={DWTNeRF: Boosting Few-shot Neural Radiance Fields via Discrete Wavelet Transform}, 
      author={Hung Nguyen and Blark Runfa Li and Truong Nguyen},
      year={2025},
      eprint={2501.12637},
      archivePrefix={arXiv},
      primaryClass={cs.CV},
      url={https://arxiv.org/abs/2501.12637}, 
}

@INPROCEEDINGS{GraspSAM,
  author={Noh, Sangjun and Kim, Jongwon and Nam, Dongwoo and Back, Seunghyeok and Kang, Raeyoung and Lee, Kyoobin},
  booktitle={2025 IEEE International Conference on Robotics and Automation (ICRA)}, 
  title={GraspSAM: When Segment Anything Model Meets Grasp Detection}, 
  year={2025},
  volume={},
  number={},
  pages={14023-14029},
  doi={10.1109/ICRA55743.2025.11128811}}

@INPROCEEDINGS{Binpicking1,
  author={Hung Nguyen, Ngoc Duy and Nguyen Nguyen, Le Hai and Pham, Phuong-Tung and Nguyen, Quoc Chi and Ly, Phuc Thanh},
  booktitle={2022 International Conference on High Performance Big Data and Intelligent Systems (HDIS)}, 
  title={Bin-Picking Solution for Industrial Robots Integrating a 2D Vision System}, 
  year={2022},
  volume={},
  number={},
  pages={266-270},
  doi={10.1109/HDIS56859.2022.9991341}}

@article{PoissonGaussian,
  author  = {B{\"a}hler, Nicolas and El Helou, Majed and Objois, {\'E}tienne and Okumu{\c{s}}, Kaan and S{\"u}sstrunk, Sabine},
  doi     = {10.1109/LSP.2022.3227522},
  journal = {IEEE Signal Processing Letters},
  number  = {},
  pages   = {2602-2606},
  title   = {PoGaIN: Poisson-Gaussian Image Noise Modeling From Paired Samples},
  volume  = {29},
  year    = {2022}
}

@inproceedings{LiteASR,
  title={LiteASR: Efficient Automatic Speech Recognition with Low-Rank Approximation},
  author={Kamahori, Keisuke and Kasai, Jungo and Kojima, Noriyuki and Kasikci, Baris},
  booktitle={Proceedings of the 2025 Conference on Empirical Methods in Natural Language Processing (EMNLP)},
  year={2025},
  pages={2047--2061},
  publisher={Association for Computational Linguistics},
  doi={10.18653/v1/2025.emnlp-main.169}
}

@INPROCEEDINGS{Harmonic,
  author={Zhou, Kaiyang and Yang, Jingkang and Loy, Chen Change and Liu, Ziwei},
  booktitle={2022 IEEE/CVF Conference on Computer Vision and Pattern Recognition (CVPR)}, 
  title={Conditional Prompt Learning for Vision-Language Models}, 
  year={2022},
  volume={},
  number={},
  pages={16795-16804},
  doi={10.1109/CVPR52688.2022.01631}}

@INPROCEEDINGS{LVIS,
  author={Gupta, Agrim and Dollár, Piotr and Girshick, Ross},
  booktitle={2019 IEEE/CVF Conference on Computer Vision and Pattern Recognition (CVPR)}, 
  title={LVIS: A Dataset for Large Vocabulary Instance Segmentation}, 
  year={2019},
  volume={},
  number={},
  pages={5351-5359},
  doi={10.1109/CVPR.2019.00550}}

@inproceedings{OpenSLR,
  author = {Tom Ko and Vijayaditya Peddinti and Daniel Povey and Sanjeev Khudanpur},
  title = {A Study on Data Augmentation of Reverberant Speech for Robust Speech Recognition},
  booktitle = {IEEE International Conference on Acoustics, Speech and Signal Processing (ICASSP)},
  year = {2017}
}

@book{SpeechBook,
  title={Speech and Language Processing},
  author={Jurafsky, Daniel and Martin, James H.},
  year={2009},
  edition={2},
  publisher={Prentice Hall}
}

@article{t-SNE,
  author = {van der Maaten, Laurens and Hinton, Geoffrey},
  journal = {Journal of Machine Learning Research},
  pages = {2579--2605},
  title = {Visualizing Data using {t-SNE} },
  url = {http://www.jmlr.org/papers/v9/vandermaaten08a.html},
  volume = 9,
  year = 2008
}

@INPROCEEDINGS{AFD-SLU,
  author={Xie, Yan and Cui, Yibo and Xie, Liang and Yin, Erwei},
  booktitle={ICASSP 2026 - 2026 IEEE International Conference on Acoustics, Speech and Signal Processing (ICASSP)}, 
  title={AFD-SLU: Adaptive Feature Distillation for Spoken Language Understanding}, 
  year={2026},
  volume={},
  number={},
  pages={19432-19436},
  doi={10.1109/ICASSP55912.2026.11462824}}

@inproceedings{BERT,
  title={BERT: Pre-training of Deep Bidirectional Transformers for Language Understanding},
  author={Jacob Devlin and Ming-Wei Chang and Kenton Lee and Kristina Toutanova},
  booktitle={North American Chapter of the Association for Computational Linguistics},
  year={2019},
  url={https://api.semanticscholar.org/CorpusID:52967399}
}

@inproceedings{Whisper,
  title={Robust Speech Recognition via Large-Scale Weak Supervision},
  author={Alec Radford and Jong Wook Kim and Tao Xu and Greg Brockman and Christine McLeavey and Ilya Sutskever},
  booktitle={International Conference on Machine Learning},
  year={2022},
  url={https://api.semanticscholar.org/CorpusID:252923993}
}

@misc{KittenTTS,
  title        = {KittenTTS: State‐of‐the‐Art Lightweight Text‐to‐Speech Model},
  author       = {KittenML},
  howpublished = {GitHub repository},
  year         = {2026},
  note         = {\url{https://github.com/KittenML/KittenTTS}},
}

@ARTICLE{SpeechSurvey,
  author={Prabhavalkar, Rohit and Hori, Takaaki and Sainath, Tara N. and Schlüter, Ralf and Watanabe, Shinji},
  journal={IEEE/ACM Transactions on Audio, Speech, and Language Processing}, 
  title={End-to-End Speech Recognition: A Survey}, 
  year={2024},
  volume={32},
  number={},
  pages={325-351},
  doi={10.1109/TASLP.2023.3328283}}

@inproceedings{Connector,
    title = "{SSR}: Alignment-Aware Modality Connector for Speech Language Models",
    author = "Tan, Weiting  and
      Inaguma, Hirofumi  and
      Dong, Ning  and
      D. Tomasello, Paden  and
      Ma, Xutai",
    editor = "Salesky, Elizabeth  and
      Federico, Marcello  and
      Anastasopoulos, Antonis",
    booktitle = "Proceedings of the 22nd International Conference on Spoken Language Translation (IWSLT 2025)",
    month = jul,
    year = "2025",
    address = "Vienna, Austria (in-person and online)",
    publisher = "Association for Computational Linguistics",
    url = "https://aclanthology.org/2025.iwslt-1.5/",
    doi = "10.18653/v1/2025.iwslt-1.5",
    pages = "56--75",
    ISBN = "979-8-89176-272-5"
}

@misc{UnitreeWebsite,
  author       = {{Unitree Robotics}},
  title        = {Unitree Robotics Official Website},
  howpublished = {\url{https://www.unitree.com/}},
  note         = {Accessed: 2026-03-03},
  year         = {n.d.}
}

@inproceedings{ALBEF,
author = {Li, Junnan and Selvaraju, Ramprasaath R. and Gotmare, Akhilesh D. and Joty, Shafiq and Xiong, Caiming and Hoi, Steven C.H.},
title = {Align before fuse: vision and language representation learning with momentum distillation},
year = {2021},
isbn = {9781713845393},
publisher = {Curran Associates Inc.},
address = {Red Hook, NY, USA},
booktitle = {Proceedings of the 35th International Conference on Neural Information Processing Systems},
articleno = {742},
numpages = {12},
series = {NIPS '21}
}

@article{AdaptiveKD,
  title={Adaptive Knowledge Distillation Between Text and Speech Pre-Trained Models},
  author={Jinjie Ni and Yukun Ma and Wen Wang and Qian Chen and Dianwen Ng and Han Lei and Trung Hieu Nguyen and Chong Zhang and Binchao Ma and E. Cambria},
  journal={ICASSP 2023 - 2023 IEEE International Conference on Acoustics, Speech and Signal Processing (ICASSP)},
  year={2023},
  pages={1-5},
  url={https://api.semanticscholar.org/CorpusID:257106768}
}

@article{SAM,
  title={Segment Anything},
  author={Kirillov, Alexander and Mintun, Eric and Ravi, Nikhila and Mao, Hanzi and Rolland, Chloe and Gustafson, Laura and Xiao, Tete and Whitehead, Spencer and Berg, Alexander C. and Lo, Wan-Yen and Doll{\'a}r, Piotr and Girshick, Ross},
  journal={arXiv:2304.02643},
  year={2023}
}

@inproceedings{GraspAnything-6DoF,
    title={Language-driven 6-dof grasp detection using negative prompt guidance},
    author={Nguyen, Toan and Vu, Minh Nhat and Huang, Baoru and Vuong, An and Vuong, Quan and Le, Ngan and Vo, Thieu and Nguyen, Anh},
    booktitle={ECCV},
    year={2024}
}

@INPROCEEDINGS{GraspAnythingpp,
  author={Vuong, An Dinh and Vu, Minh Nhat and Huang, Baoru and Nguyen, Nghia and Le, Hieu and Vo, Thieu and Nguyen, Anh},
  booktitle={2024 IEEE/CVF Conference on Computer Vision and Pattern Recognition (CVPR)}, 
  title={Language-driven Grasp Detection}, 
  year={2024},
  volume={},
  number={},
  pages={17902-17912},
  doi={10.1109/CVPR52733.2024.01695}}

@INPROCEEDINGS{GraspAnything,
  author={Vuong, A. D. and Vu, M. N. and Le, H. and Huang, B. and Binh, H. T. T. and Vo, T. and Kugi, A. and Nguyen, A.},
  booktitle={2024 IEEE International Conference on Robotics and Automation (ICRA)}, 
  title={Grasp-Anything: Large-scale Grasp Dataset from Foundation Models}, 
  year={2024},
  volume={},
  number={},
  pages={14030-14037},
  doi={10.1109/ICRA57147.2024.10611277}}

@article{PDCNet,
title = {PDCNet: A lightweight and efficient robotic grasp detection framework via Partial Convolution and knowledge distillation},
journal = {Computer Vision and Image Understanding},
volume = {259},
pages = {104441},
year = {2025},
issn = {1077-3142},
doi = {https://doi.org/10.1016/j.cviu.2025.104441},
url = {https://www.sciencedirect.com/science/article/pii/S107731422500164X},
author = {Yanshu Jiang and Yanze Fang and Liwei Deng}
}

@ARTICLE{LiteGrasp,
  author={Peng, Linpeng and Cai, Rongyao and Xiang, Jingyang and Zhu, Junyu and Liu, Weiwei and Gao, Wang and Liu, Yong},
  journal={IEEE Robotics and Automation Letters}, 
  title={LiteGrasp: A Light Robotic Grasp Detection via Semi-Supervised Knowledge Distillation}, 
  year={2024},
  volume={9},
  number={9},
  pages={7995-8002},
  doi={10.1109/LRA.2024.3436336}}

@ARTICLE{unequal,
  author={Nie, Hong and Zhao, Zhou and Chen, Lu and Lu, Zhenyu and Li, Zhuomao and Yang, Jing},
  journal={IEEE Robotics and Automation Letters}, 
  title={Smaller and Faster Robotic Grasp Detection Model via Knowledge Distillation and Unequal Feature Encoding}, 
  year={2024},
  volume={9},
  number={8},
  pages={7206-7213},
  doi={10.1109/LRA.2024.3421790}}

@misc{EUSIPCO,
      title={Learnable Multi-level Discrete Wavelet Transforms for 3D Gaussian Splatting Frequency Modulation}, 
      author={Hung Nguyen and An Le and Truong Nguyen},
      year={2026},
      eprint={2602.14199},
      archivePrefix={arXiv},
      primaryClass={eess.IV},
      url={https://arxiv.org/abs/2602.14199}, 
}

@inproceedings{WaveletGaussian,
  title     = {WaveletGaussian: Wavelet-domain Diffusion for Sparse-view 3D Gaussian Object Reconstruction},
  author    = {Hung Nguyen and Runfa Li and An Le and Truong Nguyen},
  booktitle = {Proceedings of the 2026 IEEE International Conference on Acoustics, Speech and Signal Processing (ICASSP)},
  year      = {2026},
  organization = {IEEE}
}

@inproceedings{DWTGS,
  title     = {DWTGS: Rethinking Frequency Regularization for Sparse-view 3D Gaussian Splatting},
  author    = {Hung Nguyen and Runfa Li and An Le and Truong Nguyen},
  booktitle = {Proceedings of the 2025 IEEE International Conference on Visual Communications and Image Processing (VCIP)},
  year      = {2025},
  organization = {IEEE}
}

@InProceedings{AutoOpti3DGS,
    author    = {Nguyen, Hung and Le, An and Li, Blark Runfa and Nguyen, Truong},
    title     = {From Coarse to Fine: Learnable Discrete Wavelet Transforms for Efficient 3D Gaussian Splatting},
    booktitle = {Proceedings of the IEEE/CVF International Conference on Computer Vision (ICCV) Workshops},
    month     = {October},
    year      = {2025},
    pages     = {3139-3148}
}

@inproceedings{SpeechFound1,
  title={ImageBind: One Embedding Space To Bind Them All},
  author={Girdhar, Rohit and El-Nouby, Alaaeldin and Liu, Zhuang
and Singh, Mannat and Alwala, Kalyan Vasudev and Joulin, Armand and Misra, Ishan},
  booktitle={CVPR},
  year={2023}
}

@misc{SpeechFound2,
      title={LanguageBind: Extending Video-Language Pretraining to N-modality by Language-based Semantic Alignment}, 
      author={Bin Zhu and Bin Lin and Munan Ning and Yang Yan and Jiaxi Cui and Wang HongFa and Yatian Pang and Wenhao Jiang and Junwu Zhang and Zongwei Li and Cai Wan Zhang and Zhifeng Li and Wei Liu and Li Yuan},
      year={2023},
      eprint={2310.01852},
      archivePrefix={arXiv},
      primaryClass={cs.CV}
}

@InProceedings{SpeechFound3,
  title={OneLLM: One Framework to Align All Modalities with Language},
  author={Han, Jiaming and Gong, Kaixiong and Zhang, Yiyuan and Wang, Jiaqi and Zhang, Kaipeng and Lin, Dahua and Qiao, Yu and Gao, Peng and Yue, Xiangyu},
  booktitle = {Proceedings of the IEEE/CVF Conference on Computer Vision and Pattern Recognition (CVPR)},
  year={2024}
}

@inproceedings{SpeechFound4,
  title = {Large-scale Contrastive Language-Audio Pretraining with Feature Fusion and Keyword-to-Caption Augmentation},
  author = {Wu*, Yusong and Chen*, Ke and Zhang*, Tianyu and Hui*, Yuchen and Berg-Kirkpatrick, Taylor and Dubnov, Shlomo},
  booktitle={IEEE International Conference on Acoustics, Speech and Signal Processing, ICASSP},
  year = {2023}
}

@misc{HumanText1,
      title={Bridging Language and Action: A Survey of Language-Conditioned Robot Manipulation}, 
      author={Xiangtong Yao and Hongkuan Zhou and Oier Mees and Yuan Meng and Ted Xiao and Yonatan Bisk and Jean Oh and Edward Johns and Mohit Shridhar and Dhruv Shah and Jesse Thomason and Kai Huang and Joyce Chai and Zhenshan Bing and Alois Knoll},
      year={2026},
      eprint={2312.10807},
      archivePrefix={arXiv},
      primaryClass={cs.RO},
      url={https://arxiv.org/abs/2312.10807}, 
}

@article{HumanIntro,
author = {Duncan, John A. and Alambeigi, Farshid and Pryor, Mitchell W.},
title = {A Survey of Multimodal Perception Methods for Human–Robot Interaction in Social Environments},
year = {2024},
issue_date = {December 2024},
publisher = {Association for Computing Machinery},
address = {New York, NY, USA},
volume = {13},
number = {4},
url = {https://doi.org/10.1145/3657030},
doi = {10.1145/3657030},
journal = {J. Hum.-Robot Interact.},
month = oct,
articleno = {46},
numpages = {50}
}

@InProceedings{WaveDiff,
    author    = {Phung, Hao and Dao, Quan and Tran, Anh},
    title     = {Wavelet Diffusion Models Are Fast and Scalable Image Generators},
    booktitle = {Proceedings of the IEEE/CVF Conference on Computer Vision and Pattern Recognition (CVPR)},
    month     = {June},
    year      = {2023},
    pages     = {10199-10208}
}

\end{document}